\documentclass[letterpaper]{article} 
\usepackage{aaai24}  
\usepackage{times}  
\usepackage{helvet}  
\usepackage{courier}  
\usepackage[hyphens]{url}  
\usepackage{graphicx} 
\usepackage{natbib}  
\usepackage{caption} 
\usepackage{algorithm}
\usepackage{algorithmic}

\usepackage{xcolor}
\usepackage{caption}
\usepackage{subcaption}
\graphicspath{{figs}}

\usepackage{multirow, booktabs}
\usepackage{makecell} 

\usepackage{newfloat}
\usepackage{listings}
\DeclareCaptionStyle{ruled}{labelfont=normalfont,labelsep=colon,strut=off} 
\floatstyle{ruled}
\newfloat{listing}{tb}{lst}{}
\floatname{listing}{Listing}
\title{

bbOCR: An Open-source Multi-Domain OCR Pipeline for Bengali Documents

}
\author{
    \equalcontrib Imam Mohammad Zulkarnain\textsuperscript{\rm 1},
    \equalcontrib Shayekh Bin Islam\textsuperscript{\rm 1}, 
    Md. Zami Al Zunaed Farabe\textsuperscript{\rm 1},
    \textsuperscript{\rm \#}Md. Mehedi Hasan Shawon\textsuperscript{\rm 1},
    \textsuperscript{\rm \#}Jawaril Munshad Abedin\textsuperscript{\rm 1},
    Beig Rajibul Hasan\textsuperscript{\rm 1},
    Marsia Haque Meghla\textsuperscript{\rm 2},
    Istiak Shihab\textsuperscript{\rm 2},
    Syed Mobassir\textsuperscript{\rm 2,3},
    MD. Nazmuddoha Ansary\textsuperscript{\rm 2,3},
    Asif Sushmit\textsuperscript{\rm 2},
    Farig Sadeque\textsuperscript{\rm 1,2}
}
\affiliations{
    \textsuperscript{\rm 1}Brac University, 
    \textsuperscript{\rm 2}Bengali.ai, 
    \textsuperscript{\rm 3}Apsis Solutions Limited\\

    farig.sadeque@bracu.ac.bd
}

\begin{document}

\maketitle

\begin{abstract}

Despite the existence of numerous Optical Character Recognition (OCR) tools, the lack of comprehensive open-source systems hampers the progress of document digitization in various low resource languages, including Bengali.
Low-resource languages, especially those with an alphasyllabary writing system, suffer from the lack of large-scale datasets for various document OCR components such as word-level OCR, document layout extraction, and distortion correction;  which are available as individual modules in high-resource languages. 
In this paper, we introduce Bengali.AI-BRACU-OCR (bbOCR): an open-source scalable document OCR system that can reconstruct Bengali documents into a structured searchable digitized 
format that leverages a novel Bengali text recognition model and two novel synthetic datasets. 
We present extensive component-level and system-level evaluation: both use a novel diversified evaluation dataset and comprehensive evaluation metrics. 
Our extensive evaluation suggests that our proposed solution is preferable over the current state-of-the-art Bengali OCR systems. The source codes and datasets are available here: \url{https://bengaliai.github.io/bbocr} \footnote{Authors marked with '\#' contributed equally towards writing}

\end{abstract}

\section{Introduction}

Bengali is the 6th most popular language globally, used by approximately 330 million people. Millions of Bengali Literature and historical documents reside within various organizations in different formats such as pictures, scanned pdf, or hard copies. 
Despite the advancements in multilingual and Bengali Optical Character Recognition (OCR) technologies, a comprehensive OCR pipeline for Bengali remains unrealized due to resource constraints and the complex nature of the Bengali alphasyllabary scripts.
Therefore,  a comprehensive open-source OCR system is of high demand for facilitating the digitization of Bengali documents.

    
    

\begin{figure}[htbp]
     \centering
     \begin{subfigure}[b]{0.49\columnwidth}
         \centering
         \includegraphics[width=\columnwidth, height=3cm,trim=5cm 1.5cm 5.7cm 1.5cm,clip]{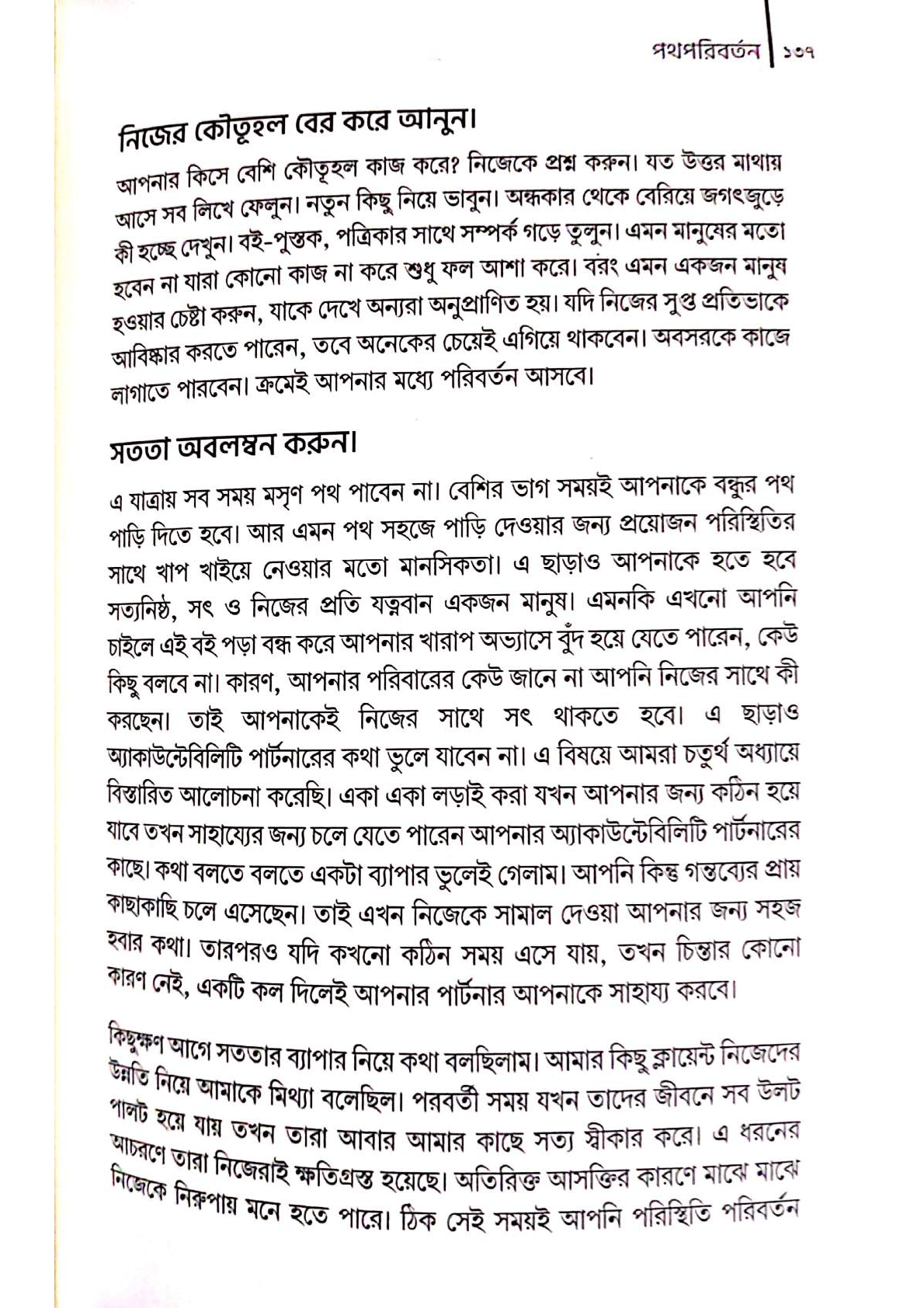}
         \caption{Scanned document}
         \label{fig:doc}
     \end{subfigure}
     \hfill
     \begin{subfigure}[b]{0.49\columnwidth}
         \centering
         \includegraphics[width=\columnwidth, height=3cm,trim=2.8cm 1.5cm 2cm 0.7cm,clip]{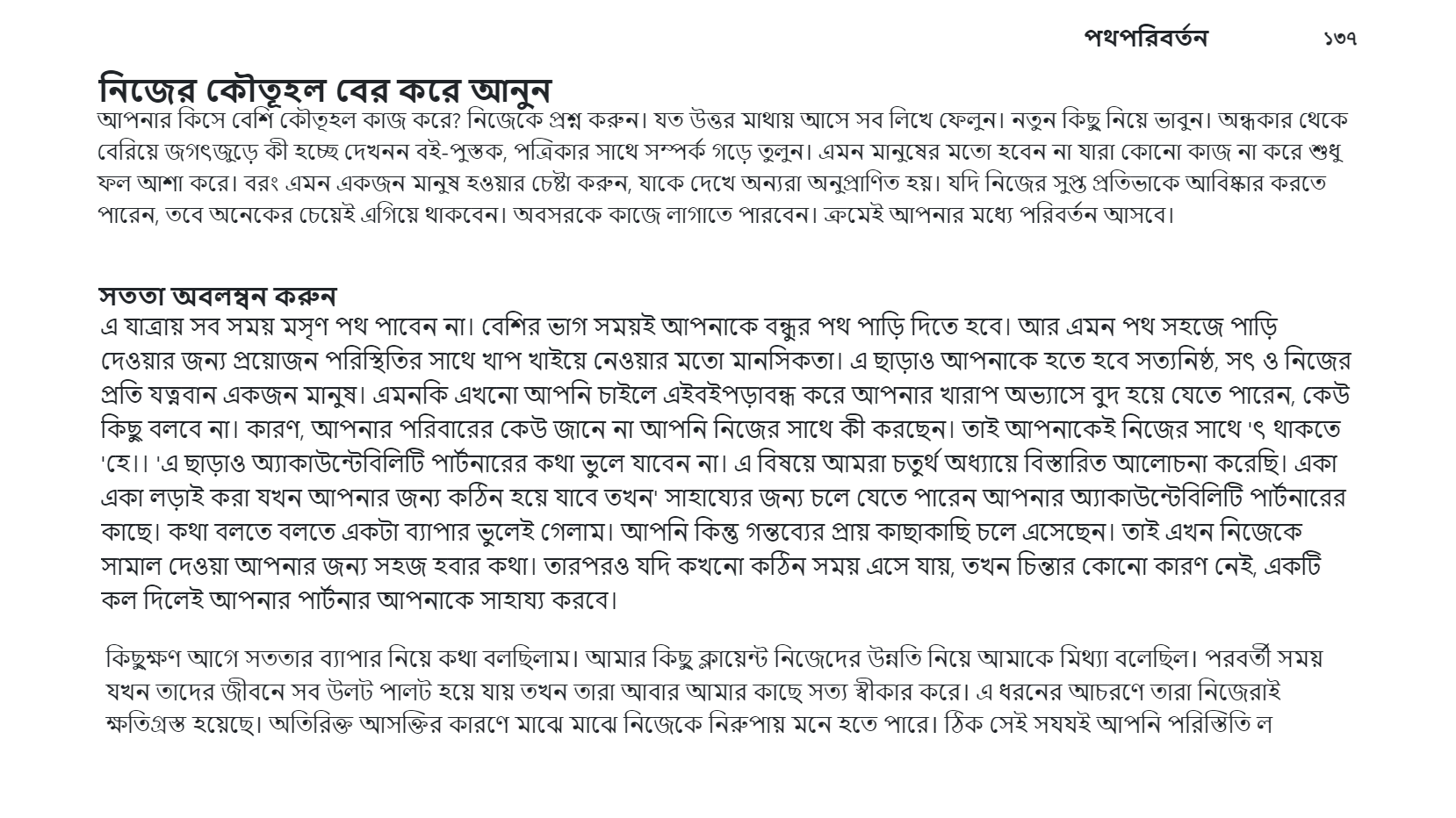}
         \caption{HTML Reconstruction}
         \label{fig:html}
     \end{subfigure}
        \caption{Illustration of proposed bbOCR system.}
        \label{fig:full-pipeline}
\end{figure}

In this paper, we present a holistic open-source Bengali OCR pipeline, Bengali.AI-BRACU-OCR (bbOCR), that is purposely built to reconstruct Bengali documents into a searchable digital format, and the illustration is shown in Figure \ref{fig:full-pipeline}. 
The whole pipeline can be broken into sub-modules. At first, the geometric correction module unwarps the scanned documents for more accurate text recognition. These unwarped images are later passed onto the illumination correction module to enhance overall image quality while discarding the shaded regions due to geometric warping. After geometric and illumination correction, the Document Layout Analysis (DLA) module is deployed to analyze the layout of the documents. This module renders four categories of region proposals inside the documents: paragraphs, text boxes, images, and tables. The region-annotated documents generated by the DLA module are transferred to the line and word detection module to first detect the lines and later the words inside each region of the documents with the help of bounding boxes. The word recognition module identifies the text characters inside the bounding boxes
and our proposed HTML reconstruction module reconstructs the original document layout into an editable text document utilizing the information regarding the text attributes extracted during the detection and recognition process. The entire pipeline is further optimized to ensure runtime efficiency compared to the state-of-the-art OCR benchmarks. Finally, we evaluate the entire system with a custom-made evaluation metric which is also proposed in this paper. 

In short, in this paper, we have:
\begin{itemize}
\item Developed the bbOCR pipeline for Bengali Document OCR,  comprised of layout Analysis, line detection, word detection, word recognition, and HTML reconstruction modules with geometric and illumination correction.
\item Introduced three new datasets: Bengali SynthTIGER and Bengali SynthINDIC for Bengali text recognition; and BCD3 (\textbf{B}engali \textbf{C}omplete \textbf{D}ocument \textbf{D}ecomposition \textbf{D}ataset) for evaluating both the component level and system level performance.
\item Introduced a Bengali text recognition model named APSIS-Net.
\item Proposed a comprehensive evaluation metric for the full system evaluation.
\end{itemize}




\section{Related Works}

{Subcomponents of our proposed comprehensive OCR framework include correction for geometry and illumination, identification of document layout, line and word detection, Word recognition, and HTML reconstruction. This literature review aims to explore the existing SOTA models, a number of evaluation metrics, and the need for a comprehensive metric to evaluate the complete bbOCR pipeline.}


{Unwarping and improving the quality of the camera images require geometric and lighting correction. To address this issue, a CNN-based stacked U-Net model is proposed by \citet{docunet}, which flattened and corrected the distorted document picture, requiring 100K synthetic 2D train data to increase the generalization capacity. The authors also proposed a model called DewarpNet \cite{dewarpnet}, which utilized 3D datasets for training. However, another architecture, PaperEdge \cite{paperedge}, is trained on combining synthetic and real document images, which has 16.2\% better performance than DewarpNet. Feng et al. recently proposed a new framework, DocTr \cite{doctr}, which uses a self-mechanism technique to capture the global context of the document image and decode the pixel-wise displacement to correct the geometric distortion. The illumination correction transformer is then used to enhance the visual quality for better OCR. 
}


{For document layout analysis, F-RCNN is used for detection \cite{F-RCNN}, which has been thoroughly trained to generate superior region recommendations. The authors combine the Region Proposal Network (RPN) and Fast R-CNN into a single network using "attention" methods to communicate their convolutional features. Nonetheless, Mask R-CNN \cite{M-RCNN} is an extension of Faster R-CNN, which appends a branch for concurrently predicting object masks alongside the existing branch for bounding box recognition. However, The cutting-edge real-time object identification and picture segmentation model YOLOv8 from Ultralytics \cite{yolov8} is renowned for its excellent speed, accuracy, adaptability, and compatibility with various hardware platforms. For document layout analysis, macro-averaged F1 score can also be employed as the evaluation metric for text-based sequence labelling tasks in document layout analysis \cite{docbank}. For end-to-end scene text detection and layout analysis tasks, the Panoptic Quality (PQ) measure is frequently utilized as the primary assessment metric \cite{long2022towards}.
}

{
Text detection and recognition is a vital part of the OCR system and to do it, attention-based models have been recently used for text detection \cite{Att, STAN}. While text detection and recognition have been optimized separately in \cite{lukas, text-wild}, an end-to-end optimization strategy has been observed to have better performance in the overall text spotting process \cite{TS1, TS2}. Standard evaluation metrics in this domain include F-score (H-mean), precision, average precision, and mAP \cite{multiplexed}. Other metrics such as Character Recognition Rate (CRR), Word Recognition Rate (WRR), and Word Accuracy can also assess the accuracy of recognized text \cite{gunna2021transfer, svtr}}.


%

{
While evaluating subtask performance offers valuable insights, there needs to be more in assessing the entire bbOCR pipeline due to the lack of consideration for interactions and dependencies between subtasks. This issue is more pronounced when comparing OCR systems utilizing diverse algorithms, which is crucial for languages like Bengali. To bridge these gaps, an encompassing evaluation metric is essential, which should evaluate the entire document layout understanding pipeline, considering text detection and recognition, object classification, document layout detection, and local accuracy metrics like Character Error Rate (CER) and Word Error Rate (WER). It should also include global metrics that assess the overall preservation of document structure and layout.
}

\section{Datasets}

Four datasets have been used in developing and testing the proposed OCR pipeline. This paper introduces two new datasets, \textbf{Bengali SynthTIGER} and \textbf{Bengali SynthINDIC}.
\begin{itemize}
    \item Novel Bengali SynthTIGER and Bengali SynthINDIC: For training word recognition models.
    \item BaDLAD: For document layout analysis Dataset \cite{shihab2023badlad}.
    \item Bengali Complete Document Decomposition Dataset(BCD3): Utilized for the complete OCR pipeline evaluation.
\end{itemize}

\subsection{BCD3}

{
To evaluate our OCR system, we developed a Bengali complete document decomposition dataset (BCD3) where 228 document image samples contain various instances of semantic entities (i.e. text boxes, paragraphs, and images). The documents are scanned copies from 9 different domains, and the Table \ref{tab:BCD3} shows the test documents' distribution. There are 88.5k words in the annotated dataset, and each has been manually annotated and validated.
}

\begin{table*}[]
\resizebox{\textwidth}{!}{%
\begin{tabular}{|c|c|c|l|cccl|c|}
\hline
\multirow{2}{*}{\textbf{Domain}} & \multirow{2}{*}{\textbf{Frequency}} & \multirow{2}{*}{\textbf{Avg. Word Count}} & \multirow{2}{*}{\textbf{Total Word}} & \multicolumn{4}{c|}{\textbf{Layout Objects}}                                         & \multirow{2}{*}{\begin{tabular}[c]{@{}c@{}}\textbf{Noise}\\ \& \textbf{Challenge}\end{tabular}} \\ \cline{5-8}
                        &                            &                                 &                             & \multicolumn{1}{c|}{P}  & \multicolumn{1}{c|}{Tx} & \multicolumn{2}{c|}{I}  &                                                                                  \\ \hline
Magazine                & 28                         & 379                              & 10617                         & \multicolumn{1}{c|}{10} & \multicolumn{1}{c|}{18} & \multicolumn{2}{c|}{3} & FT, CL, WM                                                                      \\ \hline
Single Column Book      & 70                         & 189                              & 13247                         & \multicolumn{1}{c|}{6} & \multicolumn{1}{c|}{4} & \multicolumn{2}{c|}{0} & WM, FD, DT                                                                      \\ \hline
Multi Column Book       & 43                         & 445                              & 19144                         & \multicolumn{1}{c|}{14} & \multicolumn{1}{c|}{36} & \multicolumn{2}{c|}{0} &  WM, FD, DT                                                                    \\ \hline
Government Document     & 12                         & 279                              & 3351                         & \multicolumn{1}{c|}{5} & \multicolumn{1}{c|}{30} & \multicolumn{2}{c|}{1} & N, HW, MP                                                                     \\ \hline
Two Page Book                    & 48                         & 572                              & 27497                         & \multicolumn{1}{c|}{14} & \multicolumn{1}{c|}{8} & \multicolumn{2}{c|}{0} & FD, DT, WM                                                                      \\ \hline
Old Newspaper           & 2                          & 2548                              & 5097                         & \multicolumn{1}{c|}{25} & \multicolumn{1}{c|}{9} & \multicolumn{2}{c|}{0} & FD, FT                                                                      \\ \hline
New Newspaper           & 5                          & 1692                              & 8460                         & \multicolumn{1}{c|}{26} & \multicolumn{1}{c|}{28} & \multicolumn{2}{c|}{4} & FT, SQ                                                                      \\ \hline
Property Document       & 5                          & 220                              & 1103                         & \multicolumn{1}{c|}{5} & \multicolumn{1}{c|}{14} & \multicolumn{2}{c|}{1} & HW, BR, N                                                                      \\ \hline
Book Cover              & 5                          & 13                              & 67                         & \multicolumn{1}{c|}{0} & \multicolumn{1}{c|}{7} & \multicolumn{2}{c|}{1} & FT, SQ, FD                                                                      \\ \hline

\textbf{Total}              & \textbf{228}                          & -                              & \textbf{88583}                         & \multicolumn{1}{c|}{\textbf{105}} & \multicolumn{1}{c|}{\textbf{157}} & \multicolumn{2}{c|}{\textbf{10}} & -                                                                      \\ \hline

\end{tabular}%
}
\caption{BCD3 Dataset Statistics. Layout Objects columns show the count of different objects in the respective domains. Here, P (Paragraph), Tx (Textbox), I (Image), and Tb (Table) are the objects. N = Numerals, HW = Hand Written, BR= Blurry, MP = Missing Portion, FD = Fading, DT = Dense Text, WM = Water Mark, FT = Diverse Fonts and Sizes, CL = Complex Layout, SQ = Scanning Quality}
\label{tab:BCD3}
\end{table*}


\subsection{BaDLAD}
{
BaDLAD \cite{shihab2023badlad} presents the first multidomain large Bengali Document Layout Analysis Dataset collected from multiple domains containing text with diverse layouts and orientations. This dataset  includes presenting a human-annotated dataset of 33,693 documents and the most organic dataset for Bengali DLA known thus far, featuring 710K polygon annotations across four categories: text boxes, paragraphs, images, and tables. BaDLAD uniquely spans six domains: books, government documents, wartime materials, newspapers, historical publications, and property deeds. This diversity marks BaDLAD as the pioneer in multidomain DLA datasets for Bengali.
}

\subsection{Bengali SynthTIGER}
{
The SynthTIGER \cite{synthtiger}, initially designed for English, encountered challenges when generating Bengali text recognition data due to its inability to support non-Latin languages like Bengali. This caused Bengali text to appear incorrectly because the engine failed to recognize necessary Bengali ligatures. Modifications were done by filtering the Bengali samples from the Open Super-large Crawled Aggregated Corpus (OSCAR) project and adapting the code. SynthTIGER generates visually appealing images by rendering target text using fonts, backgrounds, textures, and colour maps. Despite successful augmentation, the process was computationally intensive, taking three weeks on a high-spec machine to generate 10.2M samples with varying text lengths, 290 Bengali typefaces, and 50-95\% quality ratings.
}

\subsection{Bengali SynthINDIC}

{
SynthINDIC is an open-source tool mainly designed for crafting text recognition datasets in Indic languages. The platform offers diverse operations, including geometric manipulations like text rotation and perspective warping, background creation with mono-colour, Gaussian noise, quasi-crystal, and custom scenes, blending with 12 modes, weighted and custom multi-scale blending, and augmentation with Gaussian effects, dropout, distortions, blurs, and brightness changes. An essential addition is the text pixel negation operation, is creating challenging samples with missing pixels but retained structure for human comprehension. A dataset of 700,000 samples was generated in about three days using the same desktop computer as SynthTIGER, employing the gibberish dictionary and number-punctuation-grapheme mixed gibberish as the ground truth text to generate the difficult images with various font sizes (16 to 128) for diverse resolutions.
}

\section{System Description}
\begin{figure}[ht]
  \centering
  \includegraphics[width=0.49\textwidth, height=5cm]{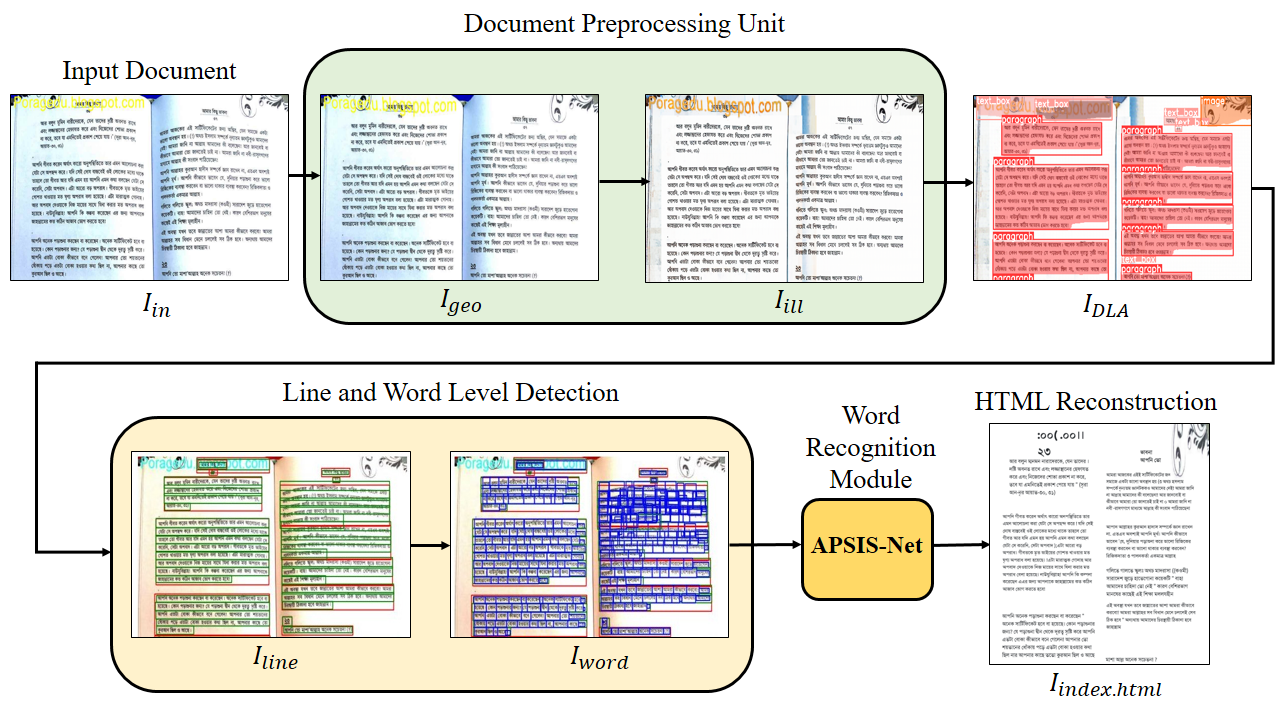}
  \caption{bbOCR system overview}
  \label{fig:system}
\end{figure}

{
We aim to produce an equivalent HTML representation of any input image containing a scanned printed page. Firstly, for an input image $I_{in}$, we apply a geometric distortion correction module to get the rectified image $I_{geo}$. Then, the geometrically corrected image is fed into the illumination correction module to get the completely corrected image $I_{ill}$. Next, the $I_{ill}$ is provided to the document layout analysis module, and we find the segmented image $I_{DLA}$. Then, the segmented image is transferred to the line and word detection module to get the line detected image $I_{line}$ and word detected image $I_{word}$ where the word detection is done using bounding boxes. Then, the $I_{word}$ image is fed to the word recognition module APSIS Network to recognize the text characters inside the bounding box and to get a recognized image. Finally, a rule-based formula converts the image into an HTML format, $I_{Index.html}$. The entire pipeline of our proposed bbOCR system is shown in Figure \ref{fig:system}.
}

\subsection{Geometric Correction}
\begin{figure}[ht]
     \centering
     \begin{subfigure}[b]{0.485\columnwidth}
         \centering
         \includegraphics[width=\columnwidth, height=3cm,trim=1.8cm 0cm 2cm 0,clip]{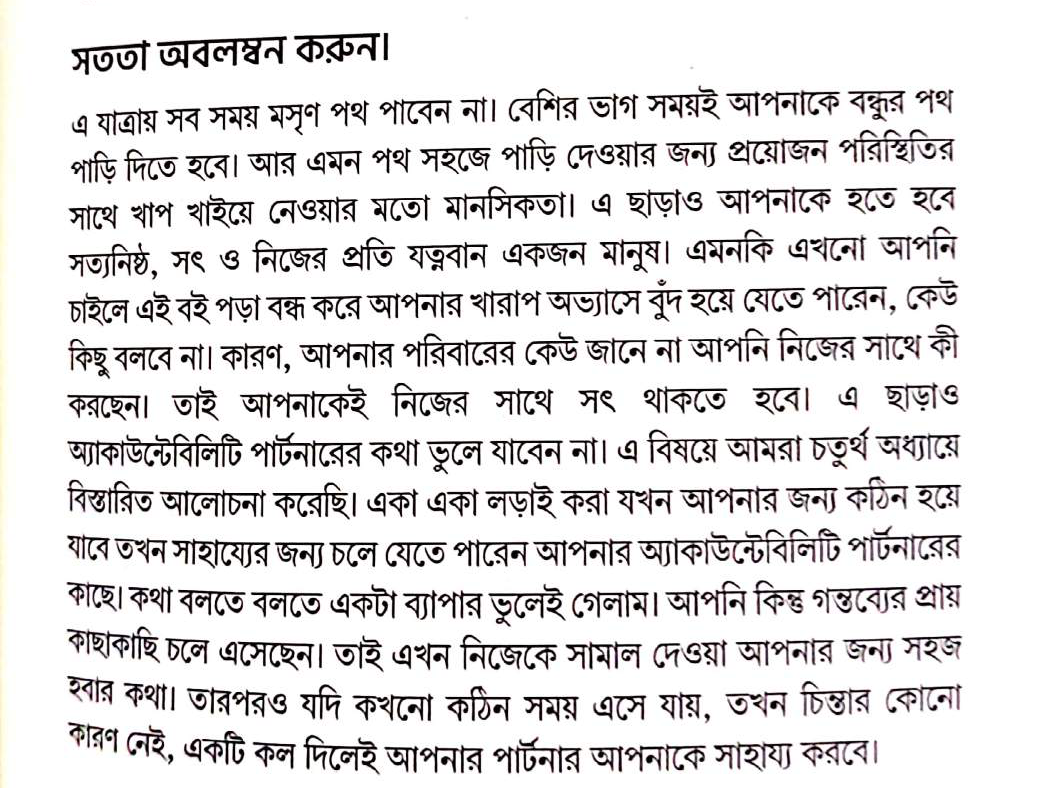}
         \caption{Scanned document before geometric correction}
         \label{fig:wo_geo}
     \end{subfigure}
     \hfill
     \begin{subfigure}[b]{0.485\columnwidth}
         \centering
         \includegraphics[width=\columnwidth, height=3cm,trim=1.6cm 0cm 1.5cm 0,clip]{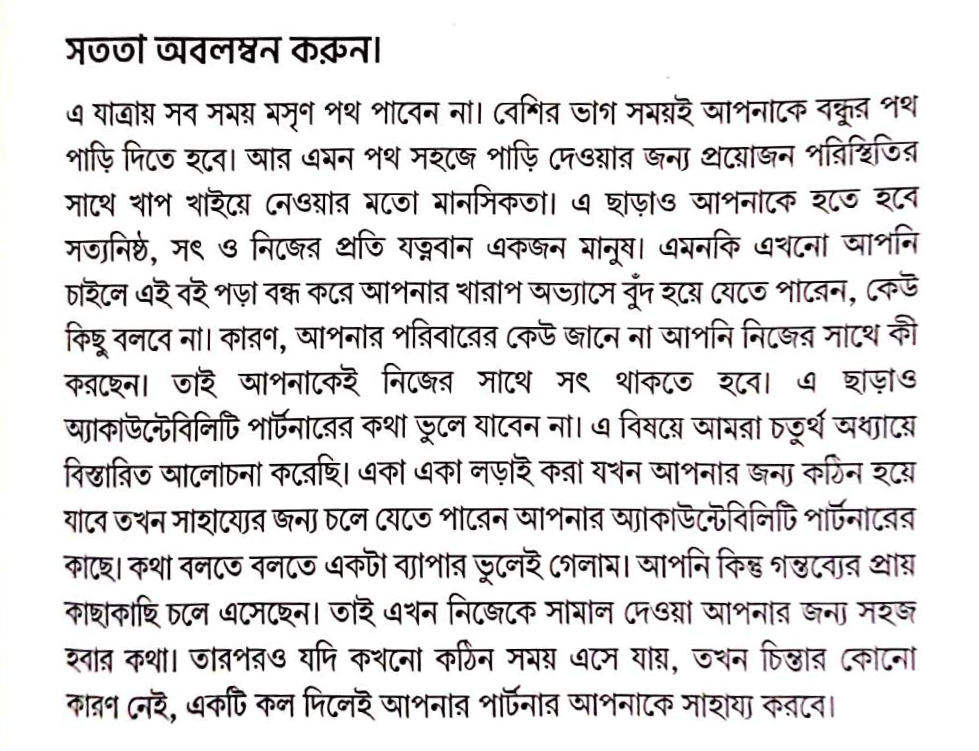}
         \caption{Scanned document after geometric correction}
         \label{fig:w_geo}
     \end{subfigure}
        \caption{Effect of geometric correction in bbOCR}
        \label{fig:geo_effect}
\end{figure}

\textcolor{black}
{The geometric and illumination correction modules are pre-processing steps to the text recognition and reconstruction modules. For geometric correction in this study, we experimented with various categories of Bengali documents on multiple models \cite{docunet, paperedge, doctr, dewarpnet}. Among those models, the PaperEdge \cite{paperedge} gives better geometric correction results than other state-of-the-art models, enhancing the recognition results in the OCR pipeline. This model also corrects cursive writing by straightening the input images. However, it could give better results for unrestricted documents for which padding has been added to the input images before applying this model. The effect of the geometric correction adopted by \cite{paperedge} can be visualized in Figure \ref{fig:geo_effect}.}

\subsection{Illumination Correction}
\begin{figure}[ht]
     \centering
     \begin{subfigure}[b]{0.49\columnwidth}
         \centering
         \includegraphics[width=\columnwidth, height = 2.5cm,trim=0.4cm 0cm 0.4cm 0,clip]{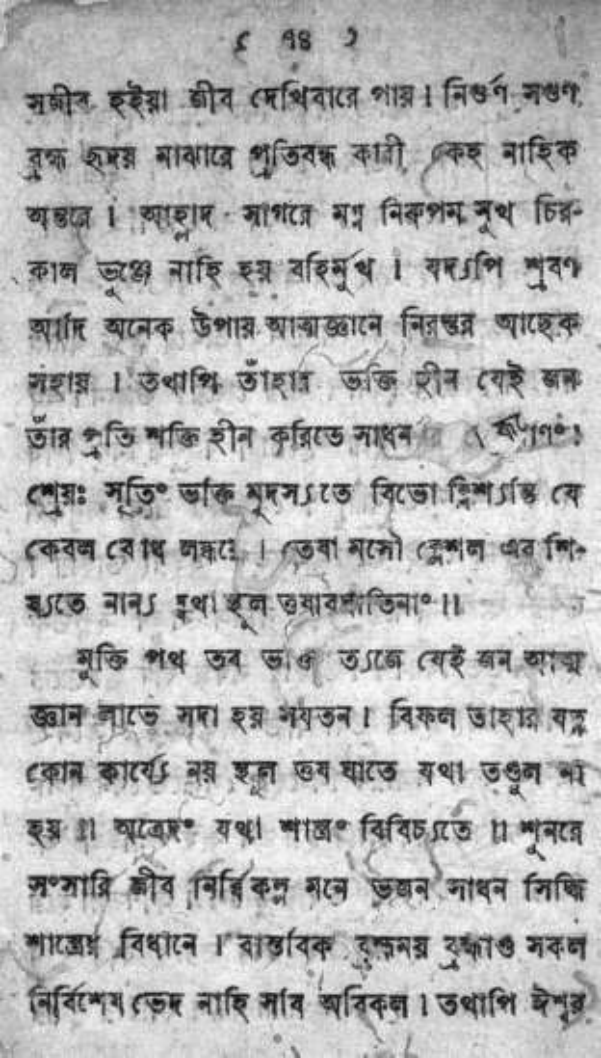}
         \caption{Scanned document before illumination correction}
         \label{fig:wo_ill}
     \end{subfigure}
     \hfill
     \begin{subfigure}[b]{0.49\columnwidth}
         \centering
         \includegraphics[width=\columnwidth, height=2.5cm,trim=0.4cm 0cm 0.4cm 0,clip]{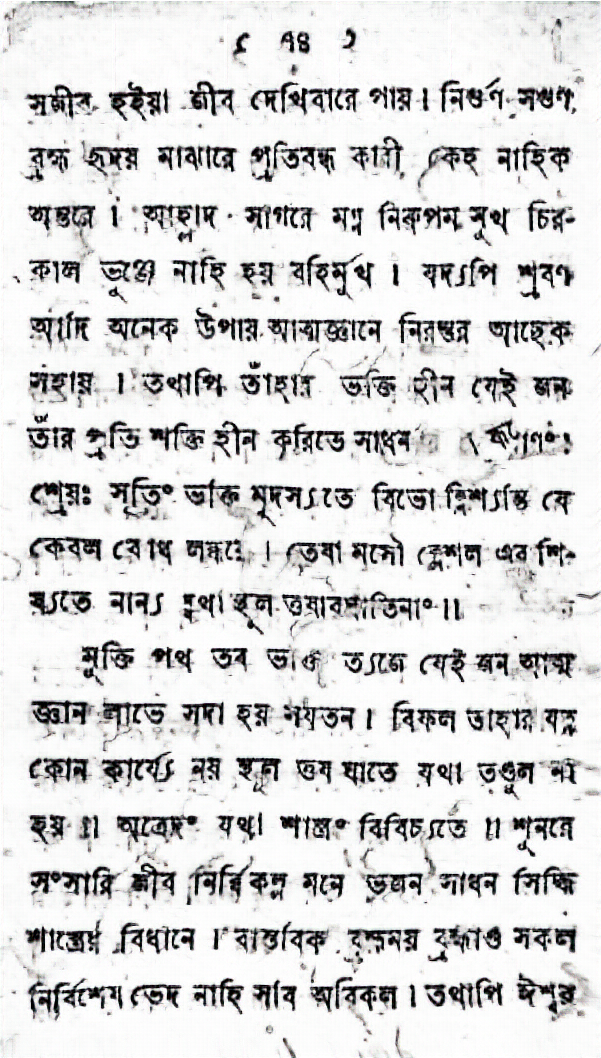}
         \caption{Scanned document after illumination correction}
         \label{fig:w_ill}
     \end{subfigure}
        \caption{Effect of illumination correction on the OCR pipeline}
        \label{fig:ill_effect}
\end{figure}

 {We are using old Bengali documents; thus, each one has a different shade of colour. Bengali letters and words are also written considerably differently and more intricately than English since, in most situations, there are additional components above or below the letters. Therefore, illumination correction is necessary, and we employed DocTR\cite{doctr} illumination correction transformation in bbOCR. We observed that OCR accuracy in our documents has significantly increased since illumination correction eliminates shading artifacts to enhance the visual quality of the documents. The effect of the illumination correction adopted by \cite{doctr} can be visualized in Figure \ref{fig:ill_effect}.}

\subsection{Document Layout Analysis}
\begin{figure}[ht]
     \centering
     \begin{subfigure}[b]{0.478\columnwidth}
         \centering
         \includegraphics[width=\columnwidth, height=4cm]{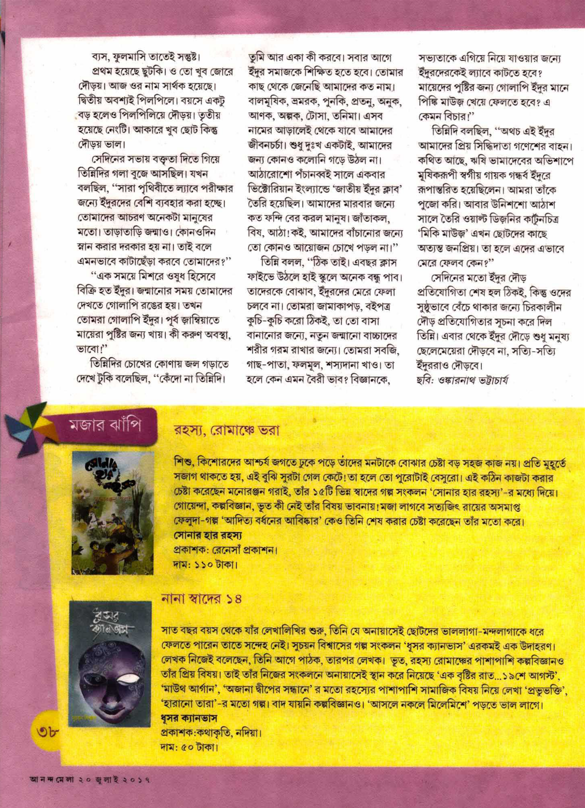}
         \caption{Input Image}
         \label{fig:input}
     \end{subfigure}
     \hfill
     \begin{subfigure}[b]{0.478\columnwidth}
         \centering
         \includegraphics[width=\columnwidth, height=4cm]{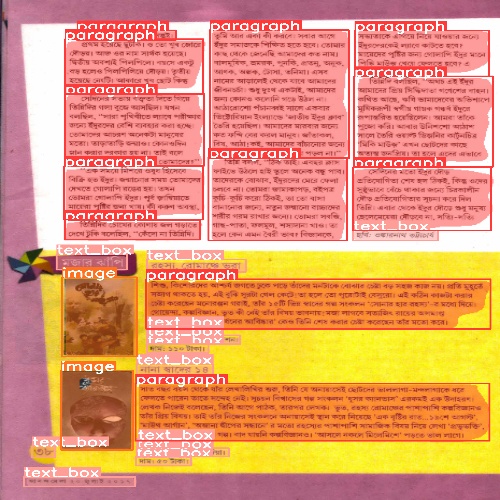}
         \caption{Output with DLA}
         \label{fig:DLA}
     \end{subfigure}
     \hfill
     
        \caption{Effect of DLA on the OCR pipeline}
        \label{fig:dla_effect}
\end{figure}

We utilized Ultralytics YOLOv8 \cite{yolov8} to analyze document layouts, the latest iteration of YOLO models, since it employs advanced deep learning techniques, and excels in tasks like detection, segmentation, pose estimation, and tracking. We trained the YOLOv8 on our BaDLAD dataset, with 16 batch sizes, 0.01 learning rate, and 100 epochs. Outputs include images, bounding boxes, masks, critical points, and labels. The outcomes of our DLA module are illustrated in Figure \ref{fig:dla_effect}. The model evaluation used mean Average Precision and Intersection over Union metrics. We assessed YOLOv8's export formats for speed-accuracy balance, finding it efficient for multi-task vision.

\subsection{Line and Word Detection}
\begin{figure}[ht]
     \centering
     \begin{subfigure}[b]{0.478\columnwidth}
         \centering
         \includegraphics[width=\columnwidth, height=4cm]{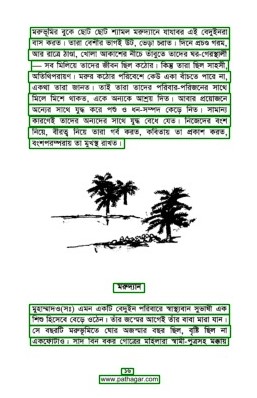}
         \caption{Line Detection}
         \label{fig:input}
     \end{subfigure}
     \hfill
     \begin{subfigure}[b]{0.478\columnwidth}
         \centering
         \includegraphics[width=\columnwidth, height=4cm]{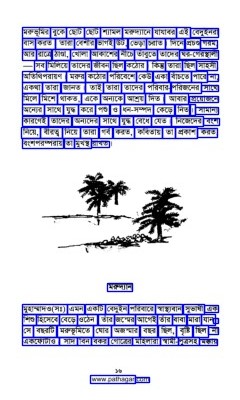}
         \caption{Word Detection}
         \label{fig:DLA}
     \end{subfigure}
     \hfill
     
        \caption{Results from the Detection module}
        \label{fig:detection}
\end{figure}

This work implements a text detection pipeline utilizing PaddlePaddle's Differentiable Binarization Network (DBNet) for line and word segmentation tasks. The DBNet framework incorporates differentiable binarization into the segmentation model, enabling adaptive thresholding and end-to-end training. For line segmentation, an English language model is deployed on DBNet with $ResNet50_vd$ and $MobileNetV3$ as the backbone, achieving state-of-the-art accuracy on benchmarks while maintaining efficiency. By leveraging DBNet in our PaddlePaddle pipeline, both line and word segmentation are performed effectively, capitalizing on DBNet's segmentation accuracy and simplified post-processing. The results from the detection module are shown in Figure \ref{fig:detection}. The modular nature of this framework allows customizing architectures and language models as required for these distinct segmentation tasks.

\subsection{Word Recognition}
\begin{figure}[ht]
  \centering
  \includegraphics[width=0.48\textwidth]{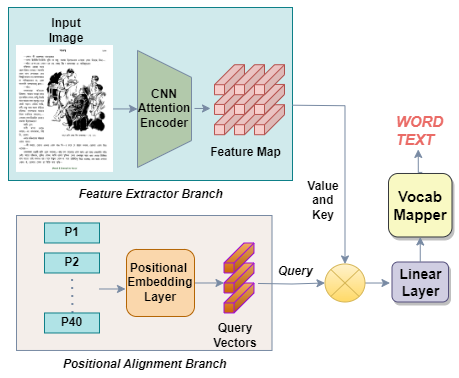}
 
  \caption{APSIS-Net Model}
  \label{fig:apsis}
\end{figure}

Following the line and word detection, we proposed a novel model called APSIS-Net consisting of a feature extractor and positional alignment branch for word-level recognition.The APSIS-Net model structure is demonstrated in Figure \ref{fig:apsis}.
The feature extractor of the APSIS-Net has a CNN Attention encoder which takes the input image of size  $H\times W\times C$ where $H, W$ and $C$ are the height, width, and channels, respectively, and transforms it into feature vectors. The encoder consists of four convolution blocks and two multi-head attention blocks, each added after two successive convolution blocks in the pipeline. Each of the convolution blocks consists of a convolutional, batch normalization, and a max pool layer. Both the height and width of the input image decrease to half the size and the number of kernels doubles after passing through each convolution block, eventually turning a $H\times W\times K$-sized image into a $\frac{H}{16} \times \frac{W}{16} \times 8K$-sized image where $K$ is the initial number of kernels. Finally the second multi-head attention block takes the $\frac{H}{16} \times \frac{W}{16} \times 8K$-sized image as input and produces the necessary key and value vectors for the positional encoding.

The positional alignment branch tries to align image feature vectors to positional values by dot-product attention where positional values are embedded and used as Query for the dot-product attention. Finally, after the positional alignment, the attention results are passed through a linear layer that creates vocabulary-sized vectors for each position. This vector is then decoded with argmax and special token handling to predict the Unicode text with a mapping module. However, this model is implemented in TensorFlow and trained on Kaggle TPU. For faster convergence, we have used torch embedding weights and passed them directly to the TensorFlow embedding layer as initial weights for the positional embedding.
\begin{table*}[ht]
    \small
    \centering
    \begin{tabular}{|r|ccc|ccc|cc|cc|}  
    
    \Xhline{1\arrayrulewidth}
        \textbf{Metric}& 
        \multicolumn{6}{c|}{\textbf{Word Level }} &
        \multicolumn{4}{c|}{\textbf{Text Level}}
        \\
        \hline
        \textbf{System}& 
        \multicolumn{3}{c|}{\textbf{Tesseract}} &
        \multicolumn{3}{c|}{\textbf{bbOCR}} &
        \multicolumn{2}{c|}{\textbf{Tesseract}} &
        \multicolumn{2}{c|}{\textbf{bbOCR}}
        \\
        \hline
        \textbf{Domain Name} & 
        \textbf{R}$ \uparrow$ & 
        \textbf{P}$ \uparrow$ & 
        \textbf{F}$ \uparrow$ & 
        \textbf{R}$ \uparrow$ & 
        \textbf{P}$ \uparrow$ & 
        \textbf{F}$ \uparrow$ & 
        \textbf{CER}$ \downarrow$ &
        \textbf{WER}$ \downarrow$ & 
        \textbf{CER}$ \downarrow$ &
        \textbf{WER}$ \downarrow$ \\
    \hline\hline
                Book & 0.18 & 0.35 & 0.21 & 0.26 & 0.43 & 0.31 & 0.86 & 1.08 & 0.50 & 0.80 \\ \hline
        Book Cover & 0.15 & 0.60 & 0.21 & 0.33 & 0.87 & 0.41 & 1.06 & 1.15 & 0.68 & 0.75 \\ \hline
        Government Document & 0.12 & 0.26 & 0.16 & 0.25 & 0.48 & 0.31 & 0.94 & 1.05 & 0.91 & 1.02 \\ \hline
        Magazine & 0.22 & 0.50 & 0.26 & 0.41 & 0.63 & 0.47 & 0.80 & 0.94 & 0.80 & 0.97 \\ \hline
        Multi Column Book & 0.27 & 0.54 & 0.32 & 0.42 & 0.62 & 0.47 & 0.85 & 1.03 & 0.64 & 0.83 \\ \hline
        New Newspaper & 0.33 & 0.69 & 0.44 & 0.29 & 0.36 & 0.32 & 0.80 & 0.92 & 0.90 & 1.04 \\ \hline
        Old Newspaper & 0.01 & 0.04 & 0.02 & 0.09 & 0.27 & 0.14 & 0.91 & 1.04 & 1.13 & 1.28 \\ \hline
        Property Document & 0.38 & 0.60 & 0.44 & 0.39 & 0.79 & 0.50 & 0.70 & 0.84 & 0.67 & 0.78 \\ \hline
        Single Column Book & 0.40 & 0.61 & 0.44 & 0.56 & 0.69 & 0.61 & 0.64 & 0.84 & 0.44 & 0.67 \\ \hline
        Per Image Average & 0.28 & 0.50 & 0.32 & \textbf{0.41} & \textbf{0.60} & \textbf{0.46} & 0.78 & 0.97 & \textbf{0.59} & \textbf{0.80} \\ \hline
    \Xhline{1\arrayrulewidth}
    	
    \end{tabular}
    \\[6pt]
    \caption{System-level performance comparison between bbOCR and Tesseract in different domains in Bengali Language.}
    \label{t1}
\end{table*}

\subsection{HTML Reconstruction}
Document Layout Analysis (DLA) is vital for converting documents into searchable HTML and determining their physical structure. Once the layout components are identified, we store their relevant data in a JSON-like format. This aids in mapping components to HTML elements. Attributes like position, dimensions, text (for paragraphs and text boxes), and image source are vital focuses. For `image' class components, we save them with their HTML source. `Paragraphs' and `text boxes', being text regions, are processed similarly. To maintain line order, lines are sorted vertically in each text region.

Afterwards, we arrange the segmented words based on their horizontal positions to preserve their sequence within each line. Subsequently, we place the identified text from each segmented word within the `text' attribute of the corresponding HTML component representing that text region. Using JSON format with attributes, we store all regions of interest. We associate each layout component with an HTML file named \texttt{index.html}, which is the final output. Notably, the `table' class is omitted due to its diverse and complex layout. When \texttt{index.html} is displayed on a web browser, paragraph text defaults to 16 pixels. To mitigate text overlap or excessive white space, dynamic font sizing adjusts text to fit within its display area, maintaining the original document's structure.

\section{Evaluation and Comparative Analysis}
Following the RRC-MLT-2019 tasks
, 
we propose precision (P), recall (R), and their harmonic mean(F) for the Word-Level (WL) reconstruction performance of the full system taking layout analysis, line detection, and word detection into consideration. We map each of the predicted text regions to the ground-truth layout components that have a maximum IOU larger than 0.5. If there is multiple non-overlapping predicted text-regions for a single ground-truth text region, we merge all the predicted words of all those predicted text regions. From this pool of predicted words, we align the each predicted word with the ground-truth words in that text region that has the maximum IOU larger than 0.5. Next, we calculate the precision, recall and F1-score from the aligned word-level ground truths and their aligned predictions.

For the Text-Level (TL) evaluation of  the bbOCR system, we compute the average of the Word Error Rate (WER) and Character Error Rate (CER) of all the paragraphs in the document weighted by their text string lengths. Similar to the word-level reconstruction metric, we first align the predicted text regions to the ground-truth layout components. During the document reconstruction, we store the text string for each predicted text region which is then compared with the paragraph level annotation to compute the aforementioned metrics.









\subsection{Runtime Performance}

To reduce the overall runtime of the entire pipeline, the Fast Deploy model of PP-OCR was adopted. Previously, an individual word from each line was sent sequentially to Paddle OCR, causing a bottleneck in line and word segmentation. To address this, multiprocessing was utilized to parallelize the line and word segmentation tasks, employing 2 processes each for line segmentation and word segmentation, respectively. Additionally, to further optimize the runtime, word crops were processed in batches of 160. These measures significantly improved the efficiency of the pipeline.

We report an average runtime on each domain for three different configurations - \textbf{Sys-1} (without any image preprocessing), \textbf{Sys-2} (with only Geometric Correction), and \textbf{Sys-3} (with Geometric Correction and Illumination).


\begin{table}[!ht]
    \centering
    \begin{tabular}{|r|r|r|r|}
    \hline
        \textbf{Domain Name}& \textbf{Sys-1} & \textbf{Sys-2} & \textbf{Sys-3} \\ \hline
        Book & 1.45 & 1.68 & 72.40 \\ \hline
        Book Cover & 0.13 & 0.17 & 12.45 \\ \hline
        Government Document & 0.51 & 0.55 & 13.51 \\ \hline
        Magazine & 0.90 & 0.99 & 26.34 \\ \hline
        Multi Column Book & 1.11 & 1.20 & 27.79 \\ \hline
        New Newspaper & 3.86 & 4.41 & \textbf{126.02} \\ \hline
        Old Newspaper & 4.97 & 5.02 & 20.60 \\ \hline
        Property Document & 0.57 & 0.70 & 36.23 \\ \hline
        Single Column Book & 0.54 & 0.62 & 25.39 \\ \hline
    \end{tabular}
    \\[6pt]
    \caption{System-level runtime comparison between bbOCR and Tesseract in different domains in Bengali Language.}
    
\end{table}




\subsection{Individual Component Performance}

In our BCD3 evaluation dataset, The Yolo-V8 model of our DLA component achieves box mAP50-95 of 0.72, 0.34, 0.63, and 0.76 in the classes `paragraph', `text\_box', `image' and `table' respectively with an overall score of 0.61. The accuracy of the text recognition model APSIS-Net is 75\% in the same dataset.

\subsection{Reconstruction Performance}
We compare bbOCR with Tesseract, a widely used open-source document OCR system, using our two proposed metrics: WL and TL. The performance of bbOCR is overall better than Tesseract. In the case of WL, our system 
consistently performs better in terms of harmonic mean (F) except for the `New Newspaper' domain.
In both systems, recall is poorer than precision. 
In the case of TL, bbOCR is better than Tesseract by a significant amount: 0.19 and 0.17 respectively in terms of CER and WER.
The geometric and illumination correction modules in bbOCR play a significant role in mitigating geometric distortions and illumination variations which aid the next modules in the full pipeline by minimizing error propagation.

\section{Conclusion}
In this work, we presented bbOCR, a complete OCR pipeline for digitization of Bengali scanned documents. To evaluate the performance of the bbOCR system, a novel BCD3 dataset was introduced which contains 9 categories of documents having 88.5K annotated words addressing numerous challenging instances. Extensive experiments were performed on BCD3 to compare bbOCR with Tessaract and we observed $8.44\%$ decrease in WER and $11.77\%$ decrease in CER for bbOCR on average across all domains. 
We also adopted different techniques e.g. multiprocessing, and efficient batching to decrease the runtime overhead of the overall system.    

Currently, our bbOCR pipeline only works on printed and scanned texts, not handwritten Bengali documents. It also gives below-par performance for blurred images during geometric and illumination testing. Moreover, the reconstruction ignores tables as we do not use any table structure detection model yet. 
Furthermore, we do not consider misaligned DLA boxes and word boxes during reconstruction. 
Besides, as there is no large-scale dataset for the Bengali language, we had to use purely synthetic data like SynthIndic. 
We plan to address these issues in the next iteration of the pipeline development, and we also plan on training the pipeline with datasets of other languages, such as English, so that we can use this pipeline in a multilingual setting.

\bibliography{aaai24.bib}
\end{document}